%% file: cvprfinal.tex
\documentclass[10pt,twocolumn,letterpaper]{article}

\usepackage[pagenumbers]{cvpr} % To force page numbers, e.g. for an arXiv version
\input{math_commands.tex}

\usepackage{graphicx}
\usepackage{amsmath}
\usepackage{amssymb}
\usepackage{booktabs}
\usepackage{multicol} 
\usepackage{caption}
\usepackage{lipsum}   % For fill text
\usepackage{stfloats}
\usepackage[pagebackref,breaklinks,colorlinks]{hyperref}

\usepackage[capitalize]{cleveref}
\crefname{section}{Sec.}{Secs.}
\Crefname{section}{Section}{Sections}
\Crefname{table}{Table}{Tables}
\crefname{table}{Tab.}{Tabs.}

\def\cvprPaperID{*****} % *** Enter the CVPR Paper ID here
\def\confName{CVPR}
\def\confYear{2023}
\graphicspath{ {./image/} }

\begin{document}

\title{N-pad : Neighboring Pixel-based Industrial Anomaly Detection}

\author{JunKyu Jang \qquad Eugene Hwang \qquad Sung-Hyuk Park\\
Korea Advanced Institute of Science and Technology (KAIST)\\
{\tt\small \{jbkjsm,hegene3686,sunghyuk.park\}@kaist.ac.kr}
}
\maketitle

%%%%%%%%% ABSTRACT
\begin{abstract}

Identifying defects in the images of industrial products has been an important task to enhance quality control and reduce maintenance costs. In recent studies, industrial anomaly detection models were developed using pre-trained networks to learn nominal representations. To employ the relative positional information of each pixel, we present \textit{\textbf{N-pad}}, a novel method for anomaly detection and segmentation in a one-class learning setting that includes the neighborhood of the target pixel for model training and evaluation. Within the model architecture, pixel-wise nominal distributions are estimated by using the features of neighboring pixels with the target pixel to allow possible marginal misalignment. Moreover, the centroids from clusters of nominal features are identified as a representative nominal set. Accordingly, anomaly scores are inferred based on the Mahalanobis distances and Euclidean distances between the target pixel and the estimated distributions or the centroid set, respectively. Thus, we have achieved state-of-the-art performance in MVTec-AD with AUROC of 99.37 for anomaly detection and 98.75 for anomaly segmentation, reducing the error by 34\% compared to the next best performing model. Experiments in various settings further validate our model. 
\end{abstract}
%%%%%%%%% BODY TEXT
\section{Introduction}

Humans have the inherent ability to recognize unusual or abnormal patterns that deviate from what is considered the norm. This trait is essential for various tasks in which inappropriate states must be detected. In particular, identifying defects in the images of industrial products is essential for enhancing quality control and reducing unnecessary maintenance costs. Therefore, artificial intelligence models for industrial anomaly detection have been developed to more precisely identify anomalous images and segments.

In the industrial field, most products are nominal with a rare occurrence of anomalous production. Here, an out-of-distribution classification is performed by training the distribution of nominal features using only a nominal dataset and by evaluating how the nominal and anomalous images in the test set deviate from the nominal distribution. Industrial anomaly detection has been challenging because some small-scale anomalous regions in products are often too small to distinguish. Moreover, anomalies in the industrial field vary from minor flaws, such as cracks, scratches, and holes, to significant irregularities, such as missing components, flips, and colors. To detect these anomalies well, various models based on autoencoders (AE), semi-supervised learning, generative adversarial networks (GAN), and normalizing flows have been developed. Recently, image representations were extracted from pre-trained models using ImageNet to learn the pixel-wise distributions of features without adaptation through transfer learning, which demonstrated state-of-the-art performances. To successfully use pre-trained models for anomaly detection, the assumption that nominal images are perfectly aligned is necessary for accurate pixel-wise distributions. In this sense, attempts have been made to disregard positional information during detection. Nevertheless, because the inherent properties of industrial products exist primarily in their unique shapes, the positional information of each pixel cannot be overlooked. 

Thus, we propose \underline{\textbf{N}}eighboring \textbf{\underline{P}}ixel-based industrial \textbf{\underline{A}}nomaly \textbf{\underline{D}}etection (\textit{N-pad}), which is the first attempt to employ the features of neighboring pixels to acquire positional information and minimize errors caused by misalignment. Here, two novel modules for weight application and feature aggregation of neighboring pixels are devised to estimate two nominal distributions by fully leveraging the features of neighboring pixels. Specifically, weights are applied to neighboring pixels according to the similarity values between the target pixel and its neighborhood, which are computed using the Bhattacharyya distance. By integrating the two estimated distributions for the computation of the final anomaly score, we achieved state-of-the-art performance in multiple classes of the industrial dataset with a \textbf{pixel-wise area under the receiver operating characteristic curve (AUROC)  of 98.75}, which is a \textbf{34\% improvement} compared to the existing state-of-the-art model . Various experiments are performed to demonstrate the robustness of the model performance.

\section{Related Works}
\subsection{General Anomaly detection}

Conventional models for anomaly detection have been developed to accurately learn the representative attributes of nominal data. In this sense, existing studies have primarily implemented reconstruction- and embedding-similarity-based methods. In terms of reconstruction-based methods, models are trained to learn features that reconstruct the original data, which identifies poorly reconstructed samples as anomalies. Accordingly, extensions based on AEs \cite{bergmann2019improving,bergmann2019mvtec,gong2019memorizing,liu2020towards,sakurada2014anomaly,deng2022anomaly} or GANs\cite{schlegl2017unsupervised, akcay2018ganomaly,pidhorskyi2018generative,sabokrou2018adversarially,venkataramanan2020attention,perera2019ocgan} have been proposed. In terms of embedding similarity-based methods, the latent features of nominal data were learned from the model to identify samples distinct from the nominal distribution as anomalies. As a reference for the nominal features, the center of constrained latent feature spaces\cite{ruff2018deep,yi2020patch,reiss2021panda,ruff2019deep}, geometric transformations\cite{golan2018deep,bergman2020classification,tack2020csi,sohn2020learning}, estimation of the probability density function using Gaussian mixture models \cite{rippel2021modeling,lee2018simple,zong2018deep}, and kernel density estimations \cite{latecki2007outlier} have been employed. Hence, distance-based metrics \cite{eskin2002geometric,roth2022towards,tsai2022multi} have been applied to assign distant samples with high anomaly scores.  

\subsection{Industrial anomaly detection}

Industrial anomaly detection has developed differently from general anomaly detection because learning the unique nominal features of an industrial object or texture is essential \cite{bergmann2019mvtec}. A recent trend in industrial anomaly detection is to use a model pre-trained on an external image dataset, such as ImageNet, to learn the distribution or features of the nominal dataset without transfer learning \cite{bergman2020deep}. One of the first successful applications was SPADE \cite{cohen2020sub}, which obtains a global feature set from the given network of the nominal data and applies a Euclidean distance-based measure of the k-nearest neighbor \cite{eskin2002geometric} to the feature set for image-wise anomaly detection. Another pre-trained network-based model, PaDiM \cite{defard2021padim} learned the distribution of local features at every pixel and obtained a pixel-wise anomaly map by computing the Mahalanobis distance between the pixel and its distribution \cite{mahalanobis1936generalized}. Similarly, PatchCore \cite{roth2022towards} proposed an algorithm for storing a subsampled coreset \cite{agarwal2005geometric} of the pre-trained features in a memory bank to obtain the patch-level distance between the coreset and a sample for detecting anomalies. In addition, attempts have been made to adapt the weights of the pre-trained model to identify the distribution of nominal data. FastFlow \cite{yu2021fastflow}, FEFM \cite{pefm}, and CFLOW-AD \cite{gudovskiy2022cflow} reported good performances by estimating the distribution of network-based features by normalizing the flow, and CFA \cite{lee2022cfa} implemented feature adaption through Coupled-hypersphere to better explain the distribution of nominal features.

However, there are some limitations in existing pre-trained feature-based models without the adaptation of pre-trained features. In particular, because  PADiM utilized only the nominal data of the target pixel location to compute its anomaly score, the scores may be overestimated if all nominal industrial images are not perfectly aligned. PatchCore was developed to disregard the positional information of the pixel because the anomaly scores were computed based on the distance from the core patch-level local features that were stored in a memory bank as a whole. Nevertheless, considering the positional information of each pixel is essential for anomaly detection. When augmentations of rotated images were included for prediction in CSI \cite{tack2020csi}, the predictive performance degraded, indicating that the change in position was not constructive for anomaly detection.

Thus, to overcome these limitations, the proposed model is devised to employ the information of neighboring pixels to estimate the nominal distribution of each pixel because the method of integrating the relationship between neighboring nodes with features has long been utilized in graph neural networks. Specifically, the similarities between the target pixel and its neighborhood are applied as weights to appropriately consider the information of neighboring pixels along with the target pixel. Consequently, we aim to design a model that is less affected by perfect image alignment but utilizes the positional information of pixels.
% \begin{figure*}
%   \centering
%   \begin{subfigure}{0.68\linewidth}
%     \fbox{\rule{0pt}{2in} \rule{.9\linewidth}{0pt}}
%     \caption{An example of a subfigure.}
%     \label{fig:short-a}
%   \end{subfigure}
%   \hfill
%   \begin{subfigure}{0.28\linewidth}
%     \fbox{\rule{0pt}{2in} \rule{.9\linewidth}{0pt}}
%     \caption{Another example of a subfigure.}
%     \label{fig:short-b}
%   \end{subfigure}
%   \caption{Example of a short caption, which should be centered.}
%   \label{fig:short}
% \end{figure*}

\section{Method}
\begin{figure*}[h]
    \begin{center}
    %\framebox[4.0in]{$\;$}
    \includegraphics[width=0.95\textwidth]{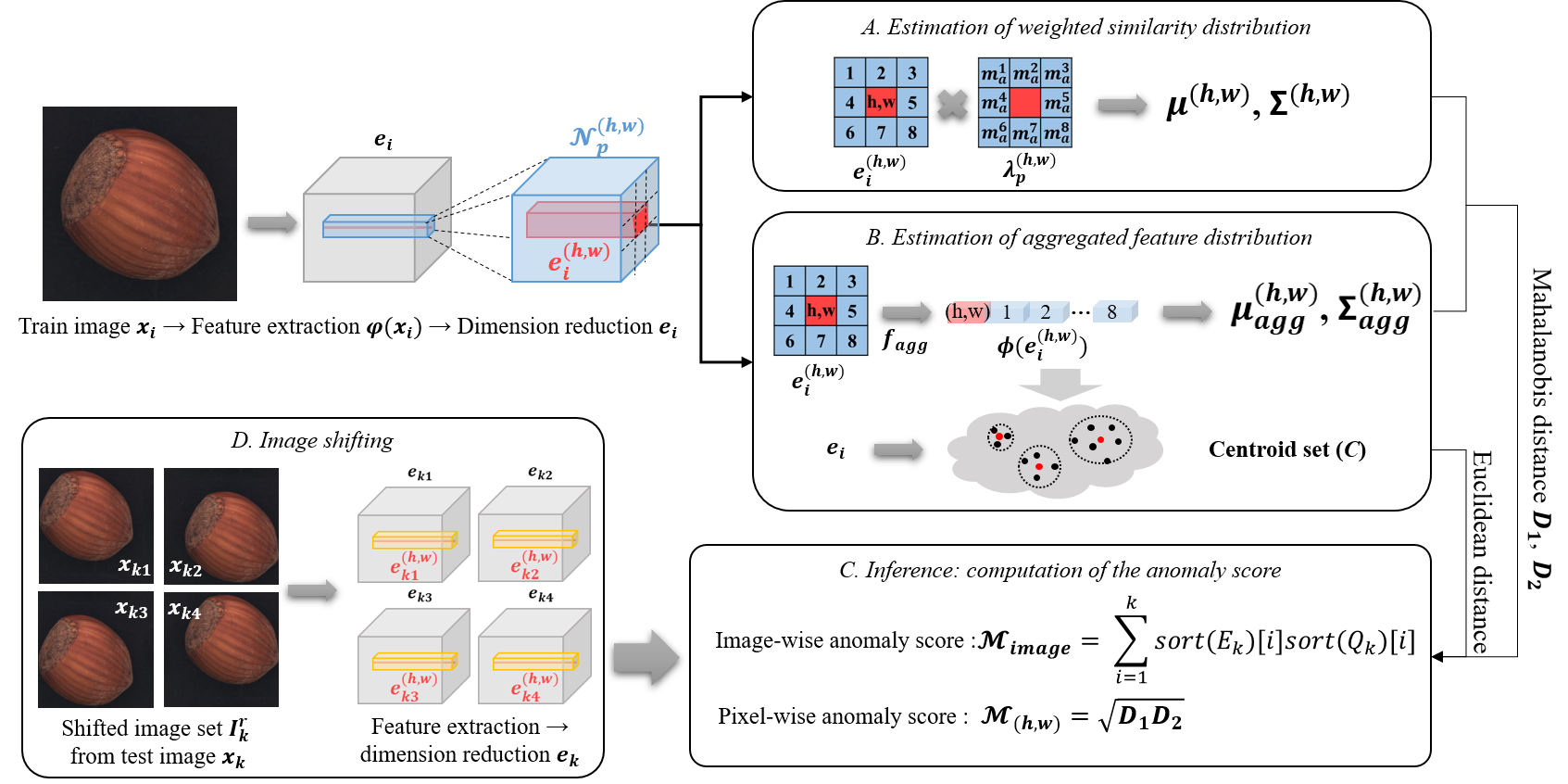}
    %\fbox{\rule[-.5cm]{0cm}{4cm} \rule[-.5cm]{4cm}{0cm}}
    \end{center}
\caption{Overall model architecture. Two nominal distributions are estimated by applying the similarity between the target pixel and its neighboring pixels as weights (A) and by aggregating features of its neighborhood (B). Also, the k-means centroids of the aggregated features are identified as a set of representative nominal features for image-level detection (B). Next, the pixel-wise anomaly map and image-wise anomaly score are computed by Mahalonobis distances between the estimated distributions and test features and Euclidean distance between the centroids of train features and test features\ (C). Lastly, a shifting technique is applied to enhance the predictive performance (D).}
\label{fig:figure1}
\end{figure*}

\subsection{Calculation of pixel-wise neighborhood similarity} 
\textbf{Feature extraction} In this study, a model architecture which implemented a pre-trained network on ImageNet as the backbone is designed for the out-of-distribution task of anomaly detection. Herein, the training set $\displaystyle \mathcal{X}_{train}=\{x_k\ |\ y_k=0\}$ consists of $\displaystyle |\mathcal{X}_{train}|=N$ nominal images, and the test set $ \displaystyle \mathcal{X}_{test}=\{x_k\ |\  y_k=0 \ or\  1\}$ consists of $|\mathcal{X}_{test}| = N_{test}$ images that are either nominal or anomalous, where $x_k$ denotes a single image from a set of all images $\mathcal{X}_{train}$, and $y_k\in\{0,1\}$ denotes image $x_k$ as nominal with 0 and anomalous with 1. As in previous studies \cite{cohen2020sub,roth2022towards,defard2021padim,bergman2020deep}, ResNet-like architectures, such as ResNet50 and WideResnet-50, were employed to extract feature maps. Within the given network $\varphi_h$, feature maps are extracted from the final output of the spatial resolution block at a specific hierarchy level $(h=1,2,3)$. Because the feature map extracted from the lowest hierarchy level $(h=1)$ has the largest size, the feature maps of higher hierarchy levels $(h=2,3)$ are interpolated to this size. Consequently, the pre-trained feature set $\varphi(x_k) = [\varphi_h(x_k)\  ,\  h = \{1,2,3\}]$ is constructed by concatenating all channels from each level.

\textbf{Dimension Reduction} Before estimating the nominal distributions, dimension reduction is performed on the total set of concatenated features $\varphi(x_k)$ because features extracted from the pre-trained network may infer redundant information. Although the concatenated channels at each pixel are assumed to follow a multivariate Gaussian distribution, not every channel may follow a Gaussian distribution. In this sense, when reducing the number of channels, we aim to select the channels with an approximate Gaussian distribution form. We believe that the normal distribution may be distorted when all channels with values below zero are set to zero after applying ReLU function at the end of most pre-trained networks. Accordingly, the nonzero values in the nominal features are counted for each channel, and the top-$d$ channels with the least nonzero values are selected. Consequently, the final nominal feature set reduced from $\varphi(x_k)$ is identified and denoted as $e_{x_k}$.
% \vspace*{-0.3cm}
\subsection{Estimation of weighted similarity distribution}
\textbf{Calculation of pixel-wise neighbor Bhattacharyya distance} To estimate the nominal distribution at each pixel, we propose a novel method for computing the pixel-wise similarity between a pixel and its neighborhood. In this study, we aim to calibrate possible misalignments by including information from neighboring pixels, whereas the perfect alignment of pixels was essential for position-based estimations in existing methods.  Specifically, the neighborhood of a pixel is defined as the set of $p$ pixels that were adjacent to the target pixel:
\begin{align}
\begin{split}
\displaystyle \mathcal{N}_p^{(h,w)} = \{(h',w')\ |\ h' &\in\ [h-\lfloor p/2\rfloor,h+\lfloor p/2 \rfloor], \\
\ w' &\in\ [w-\lfloor p/2\rfloor ,w+\lfloor p/2\rfloor] \}
\end{split}
\end{align}

First, based on the assumption that every pixel $(h, w)$ in a feature map $e_i^{\left(h,w\right)}$ of a nominal image $i$ follows a multivariate Gaussian distribution, the sample mean $\mu^{\left(h,w\right)}$ and covariance $\Sigma^{\left(h,w\right)}$ of the nominal distribution are estimated. In addition, a regularization term $\epsilon I$ is added to $\Sigma^{\left(h,w\right)}$ to ensure full rank and invertibility.     
% \begin{align}
% \displaystyle \mu^{\left(h,w\right)}=\frac{1}{N}\sum_{i = 1}^{N} e_i^{\left(h,w\right)} 
% \end{align}
% \begin{align*}
% \Sigma^{\left(h,w\right)}\!=\frac{1}{N\!\!-\!1}\sum_{i=1}^{N}\!{\left(e_i^{\left(h,w\right)}-\mu^{\left(h,w\right)}\right)\!\!\!\left(e_i^{\left(h,w\right)}-\mu^{\left(h,w\right)}\right)\!^T}\!\!\!+\epsilon I
% \end{align*}

\begin{gather}
% \begin{split}
\displaystyle \mu^{\left(h,w\right)}=\frac{1}{N}\sum_{i = 1}^{N} e_i^{\left(h,w\right)}\\ 
\Sigma^{\left(h,w\right)}\!=\frac{1}{N\!\!-\!\!1}\!\!\sum_{i=1}^{N}\!{\left(e_i^{\left(h,w\right)}-\mu^{\left(h,w\right)}\right)\!\!\!\left(e_i^{\left(h,w\right)}-\mu^{\left(h,w\right)}\right)\!^T}\!\!\!+\!\epsilon I
% \end{split}
\end{gather}

Next, the Bhattacharyya distance $m$, which indicates the distance between two probability distributions, is computed between the target pixel and all pixels within the neighborhood. Because the Bhattacharyya coefficient $BC$ measures the overlapping degree of the two distributions, the negative exponential value of the coefficient is accepted as the similarity value. Consequently, the Bhattacharyya distance set $\lambda_p^{(h,w)}$ between the estimated distribution of pixels $(h, w)$, and $\mathcal{N}_p^{(h, w)}$ is computed as follows:
% \begin{center}

\begin{equation*}
    \displaystyle \lambda_p^{(h,w)} = \{m_{a}| a \in \mathcal{N}_p^{(h,w)}\}
\end{equation*}
\vspace{-20pt}
\begin{align}
\begin{split}
&\displaystyle m_{a}=Batt(\mathcal{N}_{(\mu^{(h,w)} , \Sigma^{(h,w)})}, \mathcal{N}_{(\mu^{a} , \Sigma^{a})}), Batt = e^{-\frac{BC}{\gamma}}\\
&\displaystyle BC(\mathcal{N}(\mu_1,\Sigma_1),\mathcal{N}(\mu_2,\Sigma_2))  =\\
&\frac{1}{8}(\mu_1 - \mu_2)^T\Sigma'^{(-1)} \left(\mu_1 - \mu_2\right) + \frac{1}{2}\log(\frac{det\Sigma'}{\sqrt{det\Sigma_1 det\Sigma_2}})
\end{split}
\label{eq4}
\end{align}
where $\mu_1$ and $\mu_2$ denote a pair of mean values obtained from the estimated distributions and $\Sigma'$ denotes the average of $\Sigma_1$ and $\Sigma_2$. Herein, a balancing parameter $\gamma$ is employed to modulate the degree to which the neighboring pixels are used to estimate the distributions. A $\gamma$ of 1 is equal to the original formulation of the Bhattacharyya distance, and larger values of $\gamma$ imply that more information is used from the neighborhood. Moreover, by assuming that the inherent information of pixels within a neighborhood, denoted as the sample covariances of $(h, w)$ and $(h’, w’)$, are similar, the logarithm of the ratio of the determinant terms in \cref{eq4} is negligible. Consequently, the final similarity with the reduced computational cost is calculated as follows:
% The original Bhattacharyya distance has a $\gamma$ of 1, but we use the balancing parameter gamma to use the neighborhood pixel information properly. The smaller the $\gamma$, the more information from neighborhood pixels is used. In addition, In Eq (3), the corresponding to the logarithm of the ratio of the determinant term was negligible because the inherent information of pixels within a neighborhood indicated as the sample covariances of $(h, w)$ and $(h’, w’)$ can be considered to be similar. As a result, the final similarity with reduced computation cost was calculated as follows:
\begin{equation}
\displaystyle BC(\mathcal{N}_{(\mu_1,\Sigma_1)},\mathcal{N}_{(\mu_2,\Sigma_2)})  \simeq \frac{1}{8}(\mu_1 - \mu_2)^T\Sigma'^{(-1)} \left(\mu_1 - \mu_2\right)
\end{equation}

\textbf{Learning the normality based on similarity} As the last step for learning the nominal distribution of each pixel, we aim to accentuate the features at specific locations that may infer more relevant information about the target pixel $(h, w)$. In this sense, weights are applied to the neighboring pixels according to their similarity to the target pixel. Accordingly, the similarity values calculated within $\mathcal{N}_p^{(h, w)}$ are utilized to estimate the weighted sample mean $\vmu^{(h,w)}$ and covariance $\mSigma^{(h,w)}$ to accurately train the distribution of each pixel from the nominal images. The weighted sample mean and covariance are defined as follows:
\begin{align}
\begin{split}
\displaystyle \vmu^{(h,w)} = &\frac{1}{N}\sum_{i=1}^{N}\sum_{a \in \mathcal{N}_p^{(h,w)}} m_{a}'e_{i}^{a}, \ m_{a}' = \frac{m_{a}}{\sum_{a\in \mathcal{N}_p^{(h,w)}} m_{a}}\\
\displaystyle \mSigma^{(h,w)} = &\frac{1}{N - \sum_{a\in \mathcal{N}_p^{(h,w)}} (m_{a}')^2}\times\\
&\sum_{i=1}^{N}\sum_{a \in \mathcal{N}_p^{(h,w)}} m_{a}'(e_{i}^{a}-\vmu^{(h,w)})(e_{i}^{a}-\vmu^{(h,w)})^T
\end{split}
\end{align}
\subsection{Estimation of aggregated feature distribution}
\textbf{Learning the normality based on neighborhood aggregate features} To best use the information of neighboring pixels, the normality based on aggregating neighborhood features (B in \cref{fig:figure1}) is learned, in addition to the normality learned with weights (A in \cref{fig:figure1}). Because neighboring pixels infer unseen information from the target pixel, an anomaly map for a receptive field with higher resolution is identified by aggregating the features within a neighborhood as follows:
\vspace{1ex}
\begin{equation}
\displaystyle \phi(e^{(h,w)}) = f_{agg}(e^{a} | a \in \mathcal{N}_p^{(h,w)})
\end{equation}
where $f_{agg}$ is the aggregation function for the neighborhood $\mathcal{N}_p^{(h,w)}$. In \textit{N-pad}, we use adaptive average pooling for $f_{agg}$. Accordingly, the pixel-wise nominal distribution is learned by computing the sample mean and variance of the aggregated features at each pixel. 

Because utilizing the Euclidean distance between the test feature  and aggregated features has been effective in image-level anomaly detection in existing studies, we aim to construct a memory bank consisting of a group of essential features. In this sense, the aggregated features from the nominal set are clustered using k-means, and the features identified as the centroid of each cluster are grouped into a representative set of features denoted as $C$. In fact, the method of retrieving centroids as key features has been highly robust for outliers and noisy features within the nominal set and reported significant performance as opposed to arbitrary feature selection \cite{do2019theoretically,wang2019centroid,yuan2020defense}. Thus, within the memory bank of all cluster centroids $C$, a group of centroids near the target feature is retrieved for image-level anomaly detection.

\subsection{Inference: computation of anomaly score}
\textbf{Pixel-wise anomaly map} The anomaly score of a pixel $(h,w)$ is computed using the Mahalanobis distance between the target pixel and distributions estimated by the two modules, %$(N_1 \ and\  \mathcal{N}(\vmu_{agg}^{(h,w)},\mSigma_{agg}^{h,w}))$%.
in which features highly deviated from the nominal distributions reported higher anomaly scores. 

First, because the information of neighboring pixels at $(h, w)$ is involved in estimating the weighted distribution of $\mathcal{N}(\vmu, \mSigma)$, features extracted at $(h, w)$ affect the values in $\mathcal{N}_p^{(h,w)}$. In this sense, we also employ the distributions of neighboring pixels when computing the anomaly score of the targeted position. Accordingly, Mahalanobis distances are computed between the target feature $e^{(h,w)}$ and its neighborhood, which is defined as a collection of estimated distributions $\mathcal{P}_{q}(e^{(h,w)})$ identified from $\mathcal{N}_q^{(h,w)}$. By applying a minimum aggregation function $f$ to the set of Mahalanobis distances, $\mathcal{D}_{1}$ is obtained for each pixel and used to calculate the anomaly score. The computation of $\mathcal{D}_{1}$ proceeds as follows:
\vspace{1ex}
% Next, since the information of neighboring pixels at $(h, w)$ were involved in estimating the weighted distribution of $\mathcal{N}(\vmu, \mSigma)$, features extracted at $(h, w)$ had affected the values in $\mathcal{N}_p^{(h,w)}$. Therefore, the anomaly score of pixel $(h,w)$ was aggregated by calculating all Mahalanobis distances for the distribution collection $\mathcal{P}_{p}(e^{(h,w)})$ estimated from $\mathcal{N}_p^{(h,w)}$. As a result, the anomaly score at $(h,w)$ with $\mathcal{P}_{p}(e^{(h,w)})$ was defined as follows:
\begin{equation}
\displaystyle \mathcal{P}_{q}(e^{(h,w)}) = \{\mathcal{N}({\vmu^{a}},\mSigma^{a}) 
    \ |\ a \ \in\  \mathcal{N}_q^{(h,w)}\}
\end{equation}
\begin{align}
\begin{split}
\displaystyle &\mathcal{D}_{1}(e^{(h,w)}, \mathcal{P}_{q}(e^{(h,w)})) = \\
&f(
        \sqrt{({e^{(h,w)}}) -\vmu^{a})^{T}(\mSigma^{a})^{-1}(({e^{(h,w)}}) -\vmu^{a})} \ | 
       \  a \in  \mathcal{N}_q^{(h,w)})
\end{split}
\label{eq9}
\end{align}
\vspace{1ex}
Next, the Mahalanobis distance  $\mathcal{D}_{2}$ between aggregated features $\phi(e^{(h,w)})$ of pixel $(h, w)$ and $\mathcal{N}(\vmu_{agg}^{(h,w)},\mSigma_{agg}^{h,w})$ is defined as follows:
\begin{align}
\begin{split}
&\displaystyle \mathcal{D}_{2}(\phi({e^{(h,w)}}) , \mathcal{N}(\vmu_{agg}^{(h,w)},\mSigma_{agg}^{h,w})) = \\ 
&\sqrt{(\phi({e^{(h,w)}}) -\vmu_{agg}^{(h,w)})^{T}(\mSigma_{agg}^{(h,w)})^{-1}(\phi({e^{(h,w)}}) -\vmu_{agg}^{(h,w)}) } 
\end{split}
\label{eq10}
\end{align}
Finally, to equalize the effects of the two pixel-wise anomaly maps $\mathcal{D}_1$ and $\mathcal{D}_2$ obtained from all pixels using \cref{eq9} and \cref{eq10}, the geometric mean of the two maps is used as the final anomaly score $ \mathcal{M}_{(h,w)}$:
\begin{equation}
\mathcal{M}_{(h,w)} = \sqrt{\mathcal{D}_1\mathcal{D}_2}
\end{equation}

\textbf{Image-wise anomaly score} The image-level anomaly score based on the Euclidean distance between the aggregated features of a test image and refined set of aggregated nominal features has been effective in existing studies. Herein, a combination of Euclidean and Mahalanobis distances is employed to detect image-level anomalies more accurately. First, the top-$k$ Mahalanobis distances $\mathcal{D}_{1}$ between the target feature $e^{(h,w)}$ and estimated distribution collection of $\mathcal{P}_{q}(e^{(h,w)})$ are identified as a set $\mathcal{Q}_k$.  Next, for all pixels included in $\mathcal{Q}_k$, the minimum Euclidean distance $d$ between the aggregated feature of a pixel $\phi(e^{v})$ and the centroids in set $C$ is calculated for all pixels and denoted as set $E_k$. Consequently, the image-level anomaly score is defined as follows: 
\begin{equation}
\displaystyle \mathcal{Q}_k = {\max_{(h,w)}}{k}(\mathcal{D}_{1}(e^{(h,w)}, \mathcal{P}_{q}(e^{(h,w)})))
\end{equation}
\begin{equation}
\displaystyle E_k = \{\min_{C}{d(\phi(e^{v}), C)}\  for \  \forall \ v\  \in\   \argmax_{k}\mathcal{D}_{1}\}
\end{equation}
\begin{equation}
\displaystyle \mathcal{M}_{image} = \sum_{i=1}^{k}sort(E_k)[i]sort(\mathcal{Q}_k)[i]
\end{equation}
where $E_k$ and $\mathcal{Q}_k$ are sorted in ascending order because the sizes of the two distance values at each pixel are not in accordance. Consequently, employing both the top-$k$ Euclidean and Mahalanobis distances is demonstrated to be robust for computing image-level anomalies.

\textbf{Image-shifting}
As the final inference step, target image $x_k$ is shifted by the pixel level from size 1 to $r$ to compute the final image-wise anomaly score and pixel-wise anomaly map based on the anomaly scores of the shifted images. By aggregating the scores from all shifted images of $x_k$ denoted as a set $\mathcal{I}_{k}^r$, we expect the marginal misalignment in the images to be negligible. Set $\mathcal{I}_{k}^r$ is defined as follows:
\vspace{1ex}
\begin{align}
\begin{split}
\displaystyle \mathcal{I}_{k}^r = \{x’_k \ |&\  x’_k[h-a ,w-b] = x_k[h,w]\ \\
&\forall (a,b) \in [\lfloor -r/2,r/2\rfloor , \lfloor -r/2,r/2\rfloor]\}
\end{split}
\end{align}
\section{Experiments}
\subsection{Dataset and experimental setup}
In this study, the proposed model is trained and evaluated on MVTec Anomaly Detection dataset (MVTec-AD) \cite{bergmann2019mvtec}, which has been widely used for industrial anomaly detection tasks in existing studies. MVTec-AD consists of ten object and five texture classes with 3,629 nominal-only images for training and 1,725 nominal and anomalous images for evaluation. Moreover, we perform additional experiments Magnetic Tile Defects (MTD) in the Appendix to further validate our model. All images are center-cropped from 256 × 256 to 224 × 224 before model training and evaluation. The proposed model with a neighborhood size $p$ of 3 for model training, neighborhood size $q$ of 2 for inference, shift size $r$ of 4, balancing parameter $\gamma$ of 0.25, dimension reduction to 550, and 10\% use of the centroids from $C$ reported the best predictive performance. 

To evaluate the performance of the proposed model, the image and pixel levels of the AUROC are measured. In addition to the AUROC, the per-region-overlap score (PRO-score), which has been widely used in existing studies to measure anomaly detection performance, is measured for pixel-level anomaly segmentation \cite{bergmann2019mvtec,Bergmann_2020_CVPR}. Herein, a PRO-curve is plotted using the average rates of correctly classified pixels for all connected anomalous components, with the false positive rates set between 0 and 0.3. Accordingly, the PRO-score is computed by normalizing the area under the PRO-curve. 
\begin{table*}[h]
\centering
% \resizebox{0.9\textwidth}{!}{%
\begin{tabular}{c|ccc|cccc}
\hline
Method     & \multicolumn{3}{c|}{Normalizing   Flow Based}           & \multicolumn{4}{c}{Pre-trained   Feature Based}                     \\ \hline
Class/Model & \textbf{FastFlow}  & \textbf{PEFM}      & \textbf{CFLOW-AD}  & \textbf{SPADE} & \textbf{PaDiM} & \textbf{PatchCore} & \textit{\textbf{N-pad}}      \\ \hline
Bottle      & \textbf{100}/98.10 & \textbf{100}/98.11 & \textbf{100}/98.14 & -   /98.4      & -/98.3         & \textbf{100}/98.6  & \textbf{100/98.91} \\
Cable      & 97.58/96.98          & 98.95/96.58 & 97.41/96.70        & -/97.2     & -/96.7     & 99.4/98.5             & \textbf{99.54/98.88} \\
Capsule    & 98.52/98.84          & 91.90/97.94 & 97.69/98.64        & -/99.0     & -/98.5     & 97.8/98.9             & \textbf{99.40/98.96} \\
Carpet     & 99.15/98.95          & 100/99.00   & 99.04/98.99        & -/97.5     & -/99.1     & 98.7/\textbf{99.1}    & \textbf{99.27}/99.03 \\
Grid       & \textbf{99.68/99.24} & 96.57/98.48 & 96.24/96.76        & -/93.7     & -/97.3     & 97.9/98.7             & 98.67/98.13 \\
Hazelnut    & 97.96/97.62        & 99.89/98.78        & \textbf{100}/98.35 & -/99.1         & -/98.2         & \textbf{100}/98.7  & \textbf{100/99.03} \\
Leather     & \textbf{100}/99.41 & \textbf{100}/99.24 & \textbf{100}/99.36 & -/97.6         & -/99.2         & \textbf{100}/99.3  & \textbf{100/99.43} \\
Metalnut   & 99.51/98.36          & 99.85/96.89 & 98.92/98.32        & -/98.1     & -/97.2     & \textbf{100}/98.4     & \textbf{100/99.19}   \\
Pill       & \textbf{98.22}/97.64 & 97.51/96.67 & 96.92/98.70        & -/96.5     & -/95.7     & 96.0/97.6             & 98.00/\textbf{99.04} \\
Screw      & 86.34/98.48          & 96.43/98.93 & 83.95/97.74        & -/98.9     & -/98.5     & 97.0/\textbf{99.4}    & \textbf{97.40}/98.80 \\
Tile       & \textbf{100}/96.45   & 99.49/95.19 & \textbf{100}/97.30 & -/87.4     & -/94.1     & 98.9/95.9             & \textbf{100/97.62}   \\
Toothbrush & 89.16/97.87          & 96.38/98.28 & 92.78/98.27        & -/97.9     & -/98.8     & 99.7/98.7             & \textbf{100/99.00}   \\
Transistor & 98.58/97.07          & 97.83/96.58 & 97.38/93.15        & -/94.1     & -/98.5     & \textbf{100}/96.4     & 99.58/\textbf{98.55} \\
Wood       & \textbf{99.56}/96.23 & 99.19/95.27 & 99.30/94.80        & -/88.5     & -/94.9     & 99.0/95.1             & \textbf{99.56/97.49} \\
Zipper     & 98.55/99.04          & 98.03/98.29 & 99.03/98.38        & -/96.5     & -/98.5     & \textbf{99.5}/98.9    & 99.34/\textbf{99.16} \\ \hline
Average    & 97.52/98.03          & 98.13/97.61 & 97.24/97.57        & -/96.0     & -/97.5     & 99.0/98.1             & \textbf{99.37/98.75} \\ \hline
\end{tabular}%
% }
\caption{Image- and pixel-wise AUROC comparison of various models on MVTec-AD dataset\\}
\label{tab:table1}
\end{table*}
\begin{table*}[h]
\centering
\begin{tabular}{c|ccc|ccc|cc}
\hline
\textbf{Model}             &\textbf{FastFlow} & \textbf{PEFM} & \textbf{CFLOW-AD}  & \textbf{SPADE} & \textbf{PaDiM} & \textbf{PatchCore} & \textit{\textbf{N-pad}} \\ \hline
\multicolumn{1}{l|}{PRO-score}                 & 93.0            & 92.4               &  91.7         & 91.7           & 92.1           & 93.5      & \textbf{95.1} \\
\multicolumn{1}{l|}{Error}                &  7.0        & 7.6              &   8.3              & 8.3            & 7.9            & 6.5           & \textbf{4.9}  \\ \hline
\end{tabular}
\caption{PRO-score of different approaches on the MVTec-AD dataset}
\label{tab:table2}
\end{table*}

\subsection{Comparison with baseline methods}
\label{section4.2}

To validate the predictive performance of the proposed model, we have benchmarked methods from general anomaly detection, pre-trained feature-based models, and existing models with state-of-the-art performance on MVTec-AD dataset. Since some existing models, such as Cflow-AD and PEFM, reported ensembled results with different image resolutions or without the 224 X 224 crop, we standardize the image size of all models prior to model training for objective comparison.

\cref{tab:table1} presents the AUROC of image-level anomaly detection and pixel-level anomaly segmentation for all 15 classes of MVTec-AD. Herein, the proposed model consistently outperforms existing models with state-of-the-art-performance in both tasks. Specifically, the reduction of error for image-level detection is 37\% compared to the pre-trained feature-based model of PatchCore. Moreover, the proposed model achieves state-of-the-art performance in pixel-wise AUROC for 12 out of 15 classes with an average AUROC of 98.75 and PRO-score of 95.1 (\cref{tab:table2}), reducing the error by 34\% and 25\%, respectively, compared to the next best performing model.

We believe that the proposed method of applying similarity between the target pixel and its neighborhood as weights successfully trained the underlying relationship within the pixels. In addition, we believe that the shifting module also contributed greatly to the predictive performance by inferring the distributions of neighboring pixels when computing anomaly score. Considering that identifying ill-produced samples is highly essential and demanding in industrial fields, our experiments have demonstrated that the proposed model may be effective in industrial anomaly detection.

\subsection{Ablation study}
\textbf{Evaluation of the effectiveness of key design components}
The effectiveness of the four modules comprising the proposed model architecture (A, B, C, and D in \cref{fig:figure1}) is evaluated by removing certain modules and comparing their predictive performances. Herein, five experiments are performed, as follows.

Experiment 1: Inference only using the weighted similarity distribution of the target pixel without the aggregated feature distribution (A).

Experiment 2: Inference only using the aggregated feature distribution without the weighted similarity distributions (B).

Experiment 3: Inference using both the weighted similarity distribution of the target pixel and aggregated feature distribution (A+B).

Experiment 4: Inference using the distributions from neighboring pixels to estimate the weighted similarity distribution (A+C).

Experiment 5: Inference using the distributions from neighboring pixels to estimate the weighted similarity and aggregated feature distributions without shifting (A+B+C). 
% \usepackage{booktabs}
% \usepackage{graphicx}
% \begin{table}[]
% \centering
% \resizebox{\columnwidth}{!}{%
% \begin{tabular}{@{}l|lllllll@{}}
% \toprule
% \textbf{Experiments} &
%   \textbf{Exper1} &
%   \textbf{Exper2} &
%   \textbf{Exper3} &
%   \textbf{Exper4} &
%   \textbf{Exper5} &
%   \textit{\textbf{N-pad}} &
%   \\ \midrule
% Pixel-wise AUROC &
%   98.38 &
%   98.42 &
%   98.59 &
%   98.57 &
%   98.65 &
%   98.75 &
%   \\
% Error &
%   1.62 &
%   1.58 &
%   1.41 &
%   1.43 &
%   1.35 &
%   1.25 &
%   \\ \bottomrule
% \end{tabular}%
% }
% \caption{}
% \label{tab:my-table}
% \end{table}

% \begin{table}[h]
% \begin{center}

% \label{tab:my-table}
% \centering

% \resizebox{\columnwidth}{!}{%
% \begin{tabular}{l|llllllll}
% \hline
% \textbf{Experiments}              & \textbf{Exper1} & \textbf{Exper2} & \textbf{Exper3} & \textbf{Exper4} & \textbf{Exper5} & \textit{\textbf{N-pad}} &  \\ \hline
% \multicolumn{1}{l|}{Pixel-wiseAUROC} & 98.38           & 98.42           & 98.59           & 98.57           & 98.65           & 98.75         &  \\
% \multicolumn{1}{l|}{Error}   & 1.62            & 1.58            & 1.41            & 1.43            & 1.35            & 1.25         &  \\ \hline
% \end{tabular}%
% }
% \caption{Evaluating the effectiveness of key design components}
% \end{center}
% \end{table}
% \usepackage{graphicx}
\begin{table}[h]
\centering
\resizebox{\columnwidth}{!}{%
\begin{tabular}{c|cccccc}
\hline
\textbf{Method}                                            & \textbf{Exper1} & \textbf{Exper2} & \textbf{Exper3} & \textbf{Exper4} & \textbf{Exper5} & \textit{\textbf{N-pad}} \\ \hline
\begin{tabular}[c]{@{}c@{}}Pixel-wise\\ AUROC\end{tabular} & 98.38           & 98.45           & 98.59           & 98.57           & 98.65           & 98.75         \\
Error                                                      & 1.62            & 1.58            & 1.41            & 1.43            & 1.35            & 1.25          \\ \hline
\end{tabular}%
}
\caption{Evaluation of the effectiveness of key design components}
\label{tab:table3}
\end{table}

\begin{figure*}[ht]
  \centering
%   \captionsetup{justification=centering}
  \begin{subfigure}{0.33\textwidth}
    \includegraphics[width=\textwidth]{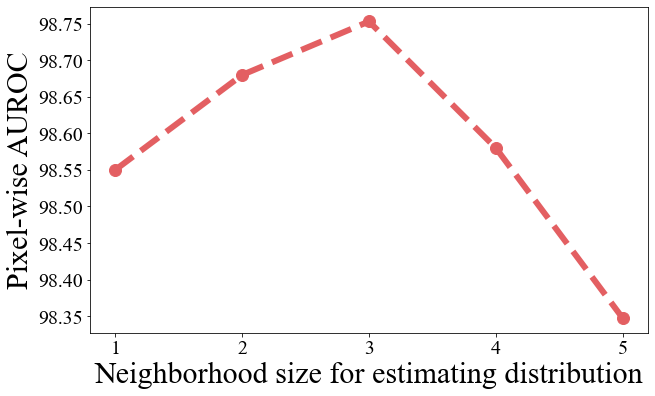}
    % \caption{An example of a subfigure.}
    \caption{AUROC comparison of different neighborhood sizes ($p$) for estimating distribution}
    \label{fig:figure2a}
  \end{subfigure}
  \hfill
   \begin{subfigure}{0.33\textwidth}
    \includegraphics[width=\textwidth]{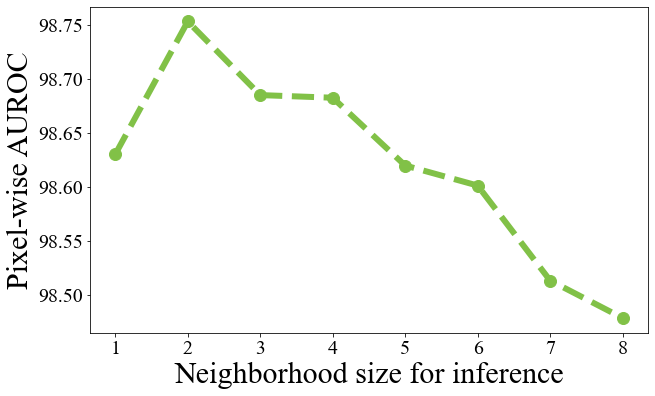}
    \caption{AUROC comparison of different neighborhood sizes ($q$) for inference}
    \label{fig:figure2b}
  \end{subfigure}
  \hfill
   \begin{subfigure}{0.33\textwidth}
    \includegraphics[width=\textwidth]{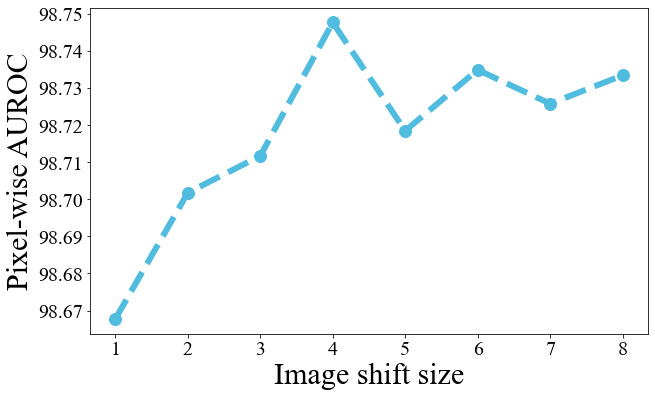}
    \caption{AUROC comparison of different image shift sizes ($r$)}
    \label{fig:figure2c}
  \end{subfigure}
   \begin{subfigure}{0.48\textwidth}
    \includegraphics[width=\textwidth]{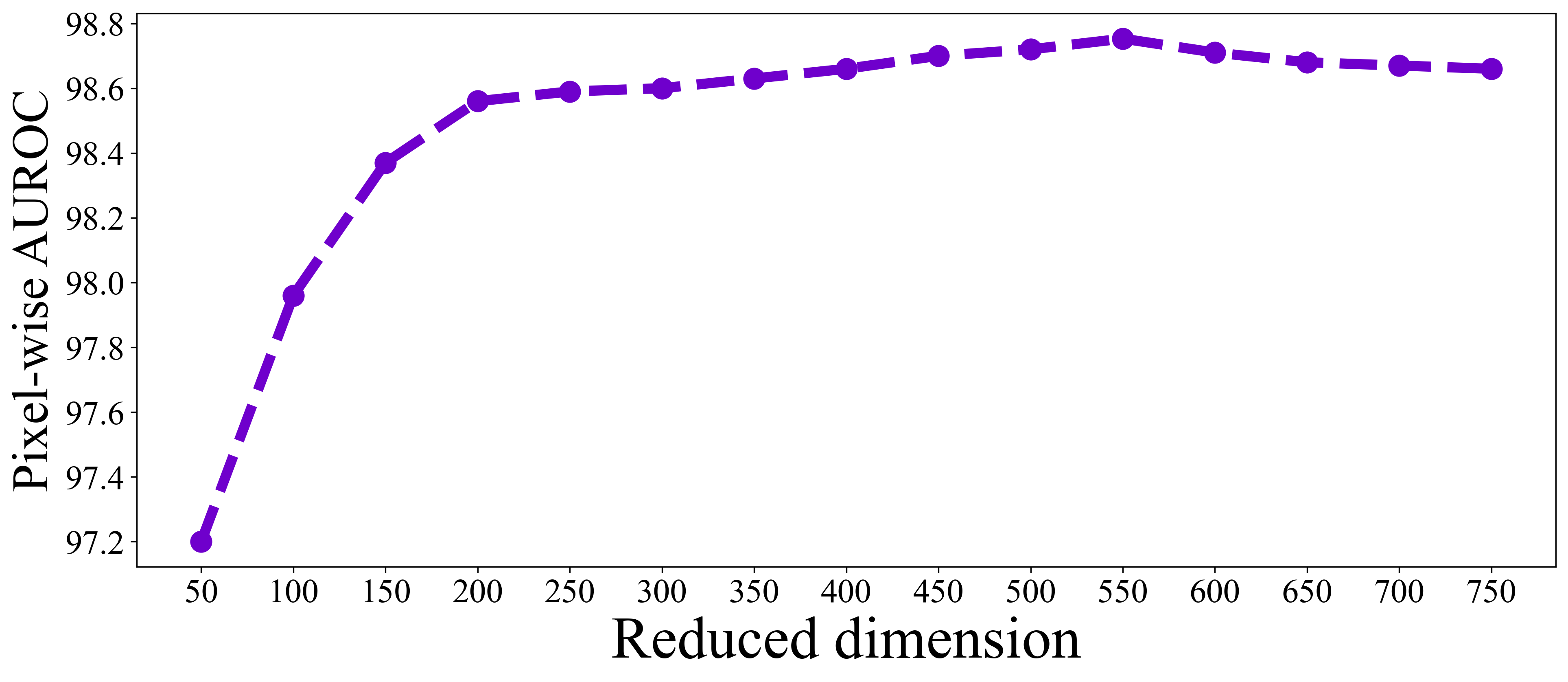}
    \caption{AUROC comparison of different dimension reduction}
    \label{fig:figure2d}
  \end{subfigure}
  \hfill
   \begin{subfigure}{0.48\textwidth}
    \includegraphics[width=\textwidth]{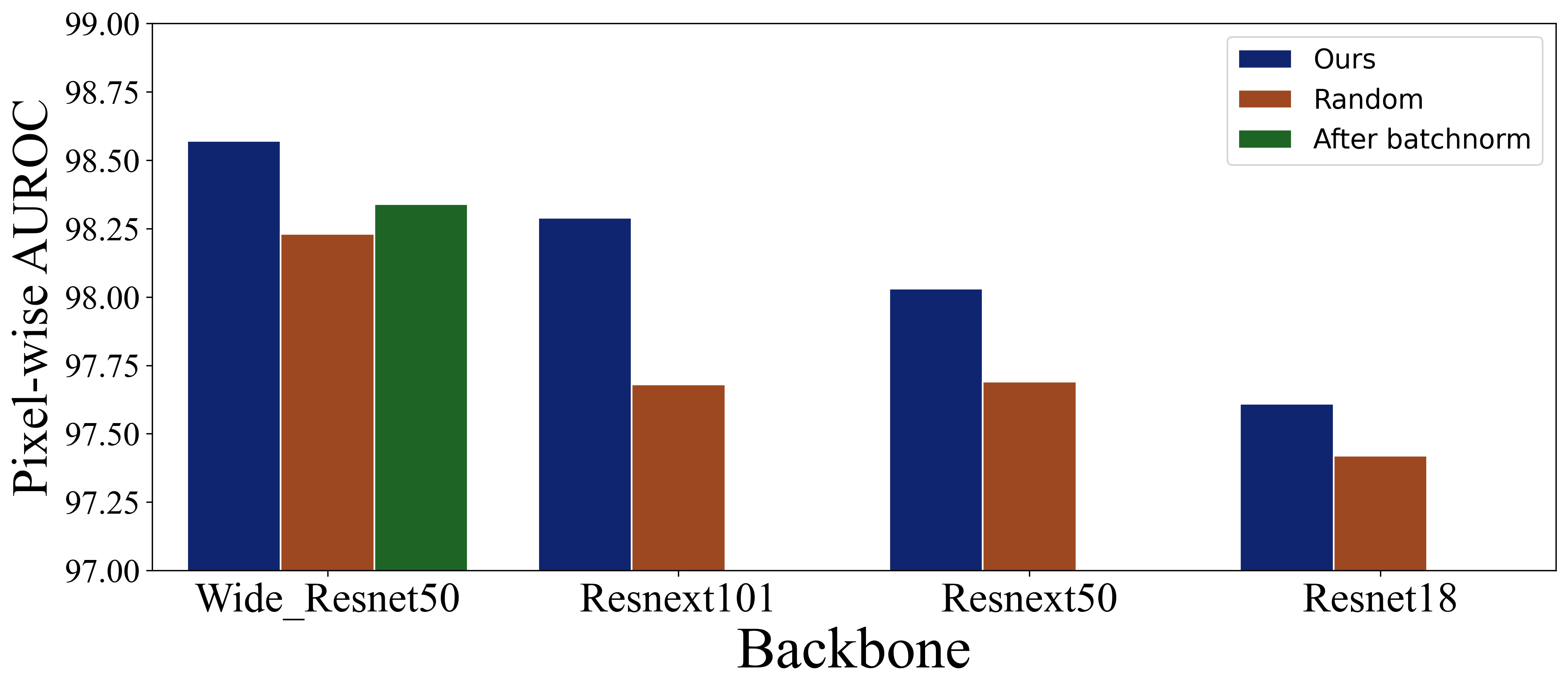}
    \caption{AUROC comparison of different backbones}
    \label{fig:figure2e}
  \end{subfigure}
\caption{Verification of parameter efficiency with various sizes and backbones.}
\label{fig:figure2}
\end{figure*}

% \begin{figure*}[ht]
% \begin{center}
% %\framebox[4.0in]{$\;$}
% \includegraphics[width=0.8\textwidth]{image/parameter.png}
% \includegraphics[width=0.8\textwidth]{image/backbonedimension.png}
% %\fbox{\rule[-.5cm]{0cm}{4cm} \rule[-.5cm]{4cm}{0cm}}
% \end{center}
% \caption{Verification of parameter efficiency with various sizes and backbones.}
% \label{fig:figure2}
% \end{figure*}
% Ablation study was performed to evaluate how the removal of certain components in the proposed model affects the predictive performance. The four main components of the model architecture were (1) learning weighted sampling based on the similarity within the neighboring pixels, (2) learning the aggregated features, (3) employing inference from neighboring pixels, and (4) shifting images. Accordingly, five experiments performed to test the effectiveness of each component were as follows:
% 1.	Utilization of aggregated features only (2)
% 2.	Inference by learning weighted sampling without utilizing distributions of neighboring pixels (1)
% 3.	Inference by learning weighted sampling with the distributions of neighboring pixels (1+3)
% 4.	Inference by learning weighted sampling with the aggregated features (1+2)
% 5.	No shifting in images (1+2+3)
\cref{tab:table3} shows that all modules significantly contribute to the performance of anomaly detection. First, the degraded performance in Experiment 5 compared with the result of \textit{N-pad} proves that shifting the aggregated anomaly maps is superior to the sole use of the original images. Next, 0.19 increase in the AUROC from Experiment 1 to Experiment 4 demonstrates that using the distributions from the neighboring pixels was effective in estimating the weighted similarity distribution. This result suggests that aggregating the anomaly scores computed from the distributions of neighboring pixels is effective. Finally, the improved performance in Experiment 3, which integrated both modules from Experiments 1 and 2, demonstrates that the inference that uses both weighted similarity and aggregated feature distributions was effective.

% \usepackage{wrapfig}

% \begin{wrapfigure}{r}{0.8\textwidth}
%   %\vspace{-20pt}
%   \begin{center}
%     \includegraphics[width=0.78\textwidth]{parameter}
%     %\includegraphics[width=0.48\textwidth]{gull}
%   \end{center}
%   %\vspace{-20pt}
%   \caption{A gull}
%   %\vspace{-10pt}
% \end{wrapfigure}

% \begin{figure}[h]
% \begin{center}
% %\framebox[4.0in]{$\;$}
% \vspace{-0.5cm}

% %\fbox{\rule[-.5cm]{0cm}{4cm} \rule[-.5cm]{4cm}{0cm}}
% \end{center}
% \vspace{-0.5cm}
% \caption{Sample figure caption.}

% \end{figure}
\textbf{Verification of parameter efficiency in model architecture}
Various parameters within the modules were tested to determine the optimal design of the proposed model. First, different neighborhood sizes ($p$) for estimating the distributions of weighted similarity or aggregated features are tested. \cref{fig:figure2a} demonstrates that the performance improves as the neighborhood size increases from 1, reaches the optimal level at a size of 3, and degrades with larger sizes. Thus, we believe that acquiring information from considerably close neighbors is the best, whereas distant neighbors infer excessive information with no greater relevance to the target pixel.

Second, the different numbers of distributions on neighboring pixels ($q$) used to compute the anomaly map in module A (\cref{fig:figure1}) are tested. \cref{fig:figure2b} shows a neighbor size of 2 as the optimal value. Because applying a large neighborhood size may affect numerous pixels of the original image during interpolation, we believe that a relatively small number is optimal.

Third, different shifting sizes ($r$) are tested, as shown in \cref{fig:figure2c}, resulting in a shifting size of 4 being the most optimal. This approach demonstrates that calibrating imperfectly aligned industrial images through the aggregation of slightly shifted versions of the images can significantly contribute to improved performance.

Fourth, the number of channels following the dimension reduction is tested by first using 50 channels and increasing the number up to 550 in units of 50. As shown in \cref{fig:figure2d}, the proposed model outperforms PatchCore, a state-of-the-art model, with an pixel-wise AUROC of 98.1, when the number reaches 150 channels, which is only 8.37\% of the total number of channels. Because the computational cost reduces quadratically with fewer channels, this result demonstrates that the proposed model can be effective with minimal computation. 
% \begin{table}[h]
% \begin{center}
% \label{tab:my-table}
% \begin{tabular}{llllll}
% \hline
% \multicolumn{1}{c}{} & \multicolumn{1}{c}{\begin{tabular}[c]{@{}c@{}}256Resize\\ \&224crop\end{tabular}} & \multicolumn{1}{c}{256Resize} & \multicolumn{1}{c}{\begin{tabular}[c]{@{}c@{}}320Resize\\ \&280Crop\end{tabular}} & \multicolumn{1}{c}{\begin{tabular}[c]{@{}c@{}}384Resize\\ \&336Crop\end{tabular}} & \multicolumn{1}{c}{Ensemble} \\ \hline
% PixelwiseAUROC       & 98.75&  98.88                             & 98.86  & 98.89   &      98.91                        \\
% \hline
% \end{tabular}
% \caption{Table 2.}
% \end{center}
% \end{table}
\begin{figure*}[ht]
  \centering
  \begin{subfigure}{0.48\linewidth}
    \includegraphics[width=\linewidth]{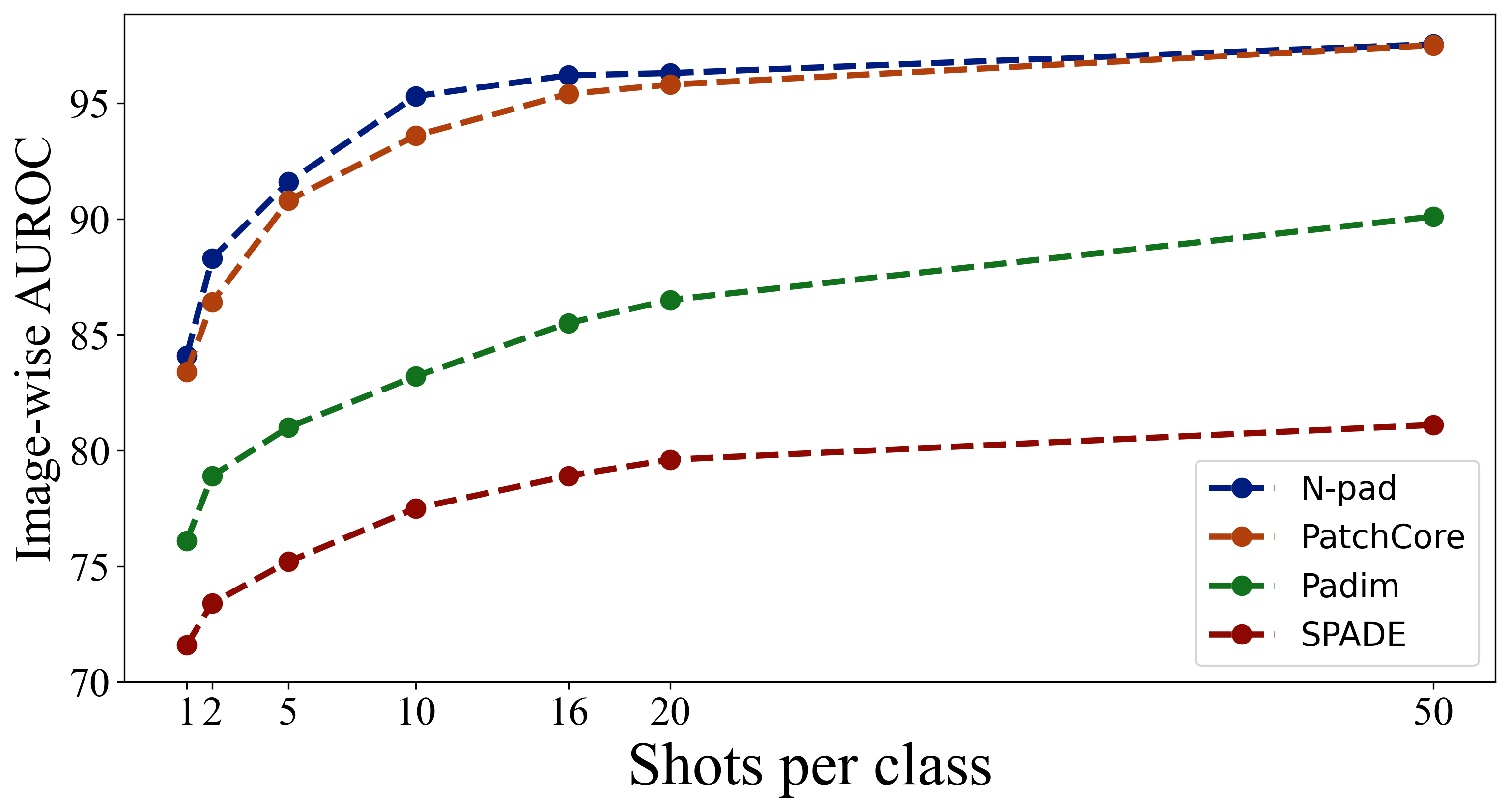}
    \caption{Image-wise AUROC comparison of different models.}
    \label{fig:figure3a}
  \end{subfigure}
  \hfill
   \begin{subfigure}{0.48\linewidth}
    \includegraphics[width=\linewidth]{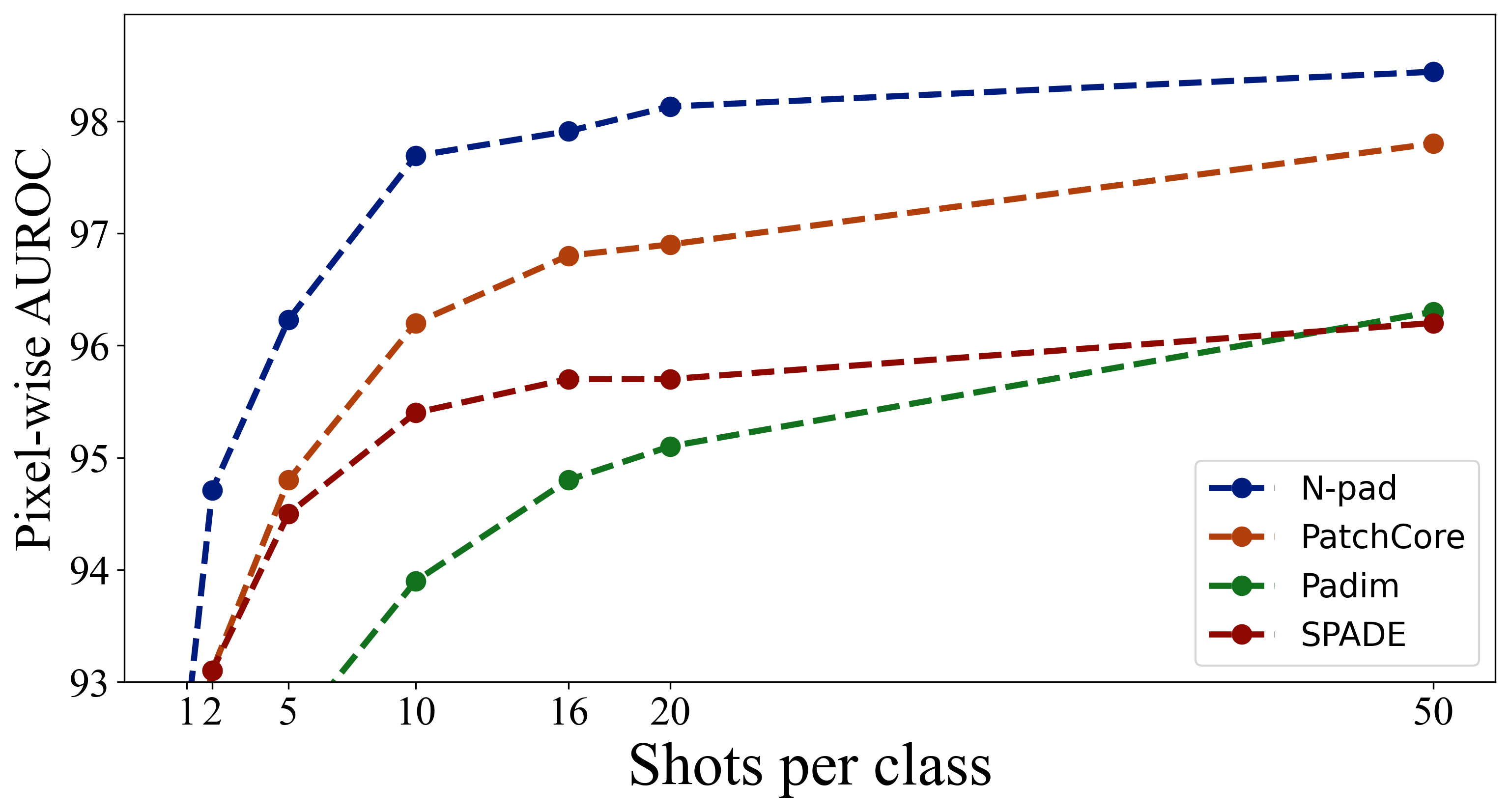}
    \caption{Pixel-wise AUROC comparison of different models.}
    \label{fig:figure3b}
  \end{subfigure}
\caption{Few shot anomaly detection performance}
\label{fig:figure3}
\end{figure*}
Finally, additional experiments are performed with different image sizes because a few existing models have reported benchmark scores by extensively reshaping the image size or excluding image crops. Because the cropped area is mostly the edge of the image background, which may be easily identified as nominal pixels, the better predictive performance is recorded with larger image sizes without cropping. Consequently, an ensemble of models which employed images cropped by 224 and 336 reported the best pixel-wise AUROC of \textbf{98.98}, as shown in \cref{tab:table4}.    
\newline
% \usepackage{graphicx}
\begin{table}[h]
\centering
\resizebox{\columnwidth}{!}{%
\begin{tabular}{c|ccccc}
\hline
\textbf{Method} &
  \textbf{\begin{tabular}[c]{@{}c@{}}256Resize\\ 224crop\end{tabular}} &
  \textbf{256Resize} &
  \textbf{\begin{tabular}[c]{@{}c@{}}320Resize\\ 280Crop\end{tabular}} &
  \textbf{\begin{tabular}[c]{@{}c@{}}384Resize\\ 336Crop\end{tabular}} &
  \textbf{Ensemble} \\ \hline
\begin{tabular}[c]{@{}c@{}}Pixel-wise\\ AUROC\end{tabular} &
  98.75 &
  98.91 &
  98.86 &
  98.89 &
  \textbf{98.98} \\ \hline
\end{tabular}%
}
\caption{AUROC comparison of different image sizes and crops}
\label{tab:table4}
\end{table}

% \begin{table}[h]
% \centering
% \resizebox{\columnwidth}{!}{%
% \begin{tabular}{c|ccccc|ccc}
% \hline
%  &
%   \multicolumn{5}{c|}{\textbf{Image Size Control}} &
%   \multicolumn{3}{c}{\textbf{Sampling Method Control}} \\ \hline
% \textbf{Method} &
%   \textbf{\begin{tabular}[c]{@{}c@{}}256Resize\\ 224crop\end{tabular}} &
%   \textbf{256Resize} &
%   \textbf{\begin{tabular}[c]{@{}c@{}}320Resize\\ 280Crop\end{tabular}} &
%   \textbf{\begin{tabular}[c]{@{}c@{}}384Resize\\ 336Crop\end{tabular}} &
%   \textbf{Ensemble} &
%   \textbf{Average} &
%   \textbf{Random} &
%   \textit{\textbf{N-pad}} \\ \hline
% \begin{tabular}[c]{@{}c@{}}Pixel-wise\\ AUROC\end{tabular} &
%   98.75 &
%   98.91 &
%   98.86 &
%   98.89 &
%   \textbf{98.98} &
%   98.42 &
%   98.42 &
%   \textbf{98.44} \\ \hline
% \end{tabular}%
% }
% \caption{Evaluation of image size efficiency and methods for estimating weighted distributions}
% \label{tab:my-table}
% \end{table}

% \begin{figure*}[h]
% \begin{center}
% %\framebox[4.0in]{$\;$}
% \includegraphics[width=\textwidth]{image/fewshot.png}
% %\fbox{\rule[-.5cm]{0cm}{4cm} \rule[-.5cm]{4cm}{0cm}}
% \end{center}
% \caption{Few shot anomaly detection performance}
% \label{fig:figure3}
% \end{figure*}

\textbf{Evaluation of the effectiveness of Bhattacharyya distance for estimating weighted similarity distributions} To demonstrate the effectiveness of the Bhattacharyya distance calculation for estimating the weighted similarity distributions in this study, uniform and random weights are tested for comparison, as presented in \cref{tab:table5}. First, weights that are randomly applied resulted in poor performance because the relationships within the pixels were not considered. Moreover, uniformly applied weights also have a minimal effect on the estimated mean and covariance of the distributions, because features from neighboring pixels may not infer significantly different information from the target pixel. Consequently, the proposed method of weighted sampling based on similarity is reported as the most effective for predictive performance.
\begin{table}[h]
\centering
\resizebox{0.6\columnwidth}{!}{%
\begin{tabular}{@{}c|ccc@{}}
\toprule
\textbf{Sampling Method}                                   & \textbf{1/n} & \textbf{Random} & \textbf{Ours} \\ \midrule
\begin{tabular}[c]{@{}c@{}}Pixel-wise AUROC\end{tabular} & 98.42        & 98.42           & \textbf{98.45}         \\ \bottomrule
\end{tabular}%
}
\caption{AUROC comparison of different sampling methods}
\label{tab:table5}
\end{table}

\textbf{Comparison of methods for reducing dimensions with various backbones}
To evaluate the proposed distribution-based method for dimension reduction, results based on a random dimension reduction with different backbones are reported for comparison. First, the random selection of dimensions in the proposed model achieves an AUROC decrease of 0.34. Next, a random dimension reduction in features extracted at the batch normalization layer prior to ReLU scores AUROC that is 0.23 lower than that of the proposed model. This result demonstrates that because the channels activated greater than 0 by the activation function are more relevant for ImageNet classification, the pre-trained network features extracted from those channels may have been more effective.  
Furthermore, various model architectures other than the WideResNet-50 of the proposed model, such as ResNet18, ResNext50, WideResNet-101, and ResNext-101, are tested for comparison. As shown in \cref{fig:figure2e}, the proposed method reports a better performance than random reduction in all architectures, demonstrating the consistency of its superiority.

\subsection{Few-shot Anomaly Detection}

\textbf{Few-shot anomaly detection}
In the industrial field, anomaly detection can be required for initial production, where only a small set of nominal sample data is available. Accordingly, few-shot anomaly detection is performed to test the proposed model with limited nominal data by testing the number of training images from 1 to 50. Consequently, the proposed model achieves better performance than the previous state-of-the-art model using only 8\% of the total dataset. Because the proposed model employs information from neighboring pixels to train the distribution of the target pixel, the augmented information from the neighborhood may have significantly contributed to few-shot learning.

\section{Conclusion}
In this paper, we propose a novel model for industrial anomaly detection and segmentation that utilizes features from the neighborhood of the target pixel. We estimate the nominal distribution of each pixel inferring the the information in the neighborhood by applying the similarity between the neighboring pixels and target pixel as weights. Moreover, another estimation of the nominal distribution based on aggregated features is proposed to employ information from various receptive fields. Various experiments evaluate the model in multiple settings and achieved state-of-the-art performances on the 15 classes of an industrial anomaly dataset. Thus, we believe that learning nominal distributions with the pre-trained features of neighboring pixels is useful and effective for improving predictive performances in industrial anomaly detection.

\appendix

%Appendix
\section*{Appendix}
\section{Evalution on MTD dataset}
In addition to the MVTec-AD dataset, we performed additional experiments with MTD (Magnetic Tile Defects) dataset which has also been used for industrial anomaly detection in previous studies\cite{huang2020surface}. MTD dataset consists of magnetic tile images in various shapes and patterns, of which 925 are nominal and 392 are anomalous. As in previous studies, 80\% of the nominal data were employed for model training and the remaining 20\% and the anomalous data were employed for evaluation. Accordingly, the results were compared to existing pre-trained network-based models and DifferNet\cite{rudolph2021same}, which reported good performance. The results are reported as follows:

\begin{table}[h]
\centering
\resizebox{\columnwidth}{!}{%
\begin{tabular}{l|llll}
\hline
\textbf{Model}               & \textbf{DifferNet} & \textbf{PaDiM} & \textbf{PatchCore} & \textit{\textbf{N-pad}}  \\ \hline
Image-wise AUROC          & 97.7               & 86.88          & 97.9               & \textbf{98.22} \\
Pixel-wise AUROC &                    & 82.45          & 84.90              & \textbf{85.43} \\ \hline
\end{tabular}%
}
\caption{Evalution on MTD dataset}
\label{tab:table6}
\end{table}

Since the shapes of nominal data are not consistent, the images may not be clearly aligned, which makes the predictions on MTD dataset challenging. Nevertheless, we have achieved superior performance by employing neighboring pixels to calculate anomaly score.

\section{Evaluation on various ratios of k-means centroids}
We have tested various ratios of the K-means centroids included for model evaluation and compared their AUROCs. In fact, the decrease in the number of clusters did not significantly affect the results because a highly robust set of centroids was employed.
% \usepackage{graphicx}
\begin{table}[h]
\centering
\resizebox{\columnwidth}{!}{%
\begin{tabular}{c|ccccc}
\hline
K-means ratio & 0.25  & 0.1   & 0.05  & 0.01  & 0.005 \\ \hline
Image-wise AUROC         & 99.39 & 99.37 & 99.34 & 99.31 & 99.10 \\ \hline
\end{tabular}%
}
\caption{Evaluation of image-wise AUROC on various ratios of k-means centroids}
\label{tab:table7}
\end{table}
\begin{table*}[!b]
\centering
\resizebox{\textwidth}{!}{%
\begin{tabular}{c|ccccccccccccccc|c}
\hline
\textbf{Model/Class} &
  \textbf{Bottle} &
  \textbf{Cable} &
  \textbf{Capsu.} &
  \textbf{Carpet} &
  \textbf{Grid} &
  \textbf{Hazel.} &
  \textbf{Leat.} &
  \textbf{Metal.} &
  \textbf{Pill} &
  \textbf{Screw} &
  \textbf{Tile} &
  \textbf{Tooth.} &
  \textbf{Trans.} &
  \textbf{Wood} &
  \textbf{Zip.} &
  \textbf{Average} \\ \hline
\textbf{FastFlow} &
  91.9 &
  89.6 &
  92.7 &
  96.3 &
  \textbf{97.4} &
  94.5 &
  \textbf{99.1} &
  93.4 &
  92.4 &
  92.6 &
  89.1 &
  83.6 &
  91.7 &
  93.0 &
  96.7 &
  93.0 \\
\textbf{CFLOW} &
  93.2 &
  92.6 &
  93.9 &
  95.3 &
  89.5 &
  95.3 &
  98.5 &
  90.2 &
  94.4 &
  91.7 &
  86.8 &
  85.7 &
  84.7 &
  90.4 &
  93.1 &
  91.7 \\
\textbf{PEFM} &
  95.4 &
  93.7 &
  93.4 &
  96.3 &
  94.8 &
  95.5 &
  98.3 &
  93.1 &
  95.2 &
  94.7 &
  81.5 &
  89.0 &
  79.9 &
  90.3 &
  95.1 &
  92.4 \\
\textbf{SPADE} &
  95.5 &
  90.9 &
  93.7 &
  94.7 &
  86.7 &
  95.4 &
  97.2 &
  94.4 &
  94.6 &
  \textbf{96.0} &
  75.6 &
  93.5 &
  87.4 &
  87.4 &
  92.6 &
  91.7 \\
\textbf{PaDIM} &
  94.8 &
  88.8 &
  93.5 &
  96.2 &
  94.6 &
  92.6 &
  97.8 &
  85.6 &
  92.7 &
  94.4 &
  86.0 &
  93.1 &
  84.5 &
  91.1 &
  95.9 &
  92.1 \\
\textbf{PatchCore} &
  96.1 &
  92.6 &
  95.5 &
  \textbf{96.6} &
  95.9 &
  93.9 &
  98.9 &
  91.3 &
  94.1 &
  97.9 &
  87.4 &
  91.4 &
  83.5 &
  89.6 &
  \textbf{97.1} &
  93.5 \\ \hline
\textit{\textbf{N-pad}} &
  \textbf{96.3} &
  \textbf{97.2} &
  \textbf{95.7} &
  95.6 &
  94.2 &
  \textbf{95.6} &
  97.2 &
  \textbf{95.1} &
  \textbf{97.1} &
  94.9 &
  \textbf{89.8} &
  \textbf{93.7} &
  \textbf{94.5} &
  \textbf{93.4} &
  \textbf{97.1} &
  \textbf{95.1} \\ \hline
\end{tabular}%
}
\caption{PRO Score comparison of various models on MVTec-AD dataset}
\label{tab:table8}
\end{table*}
\section{Details in PRO-score}
PRO-scores of each class in the MVTec-AD, which have not been listed in detail in \cref{section4.2}, are reported in \cref{tab:table8}. Here, N-pad reported the best performance in terms of PRO score for 11 out of 15 classes of MVTec-AD. Specifically, previous models reported higher scores only for carpet, grid, screw, and zipper. Thus, we may suggest that the proposed model has been developed with superior performance.
% \usepackage{graphicx}

\section{Visual result}

\begin{figure*}[ht]
\begin{center}
%\framebox[4.0in]{$\;$}
\includegraphics[width = 0.95\textwidth]{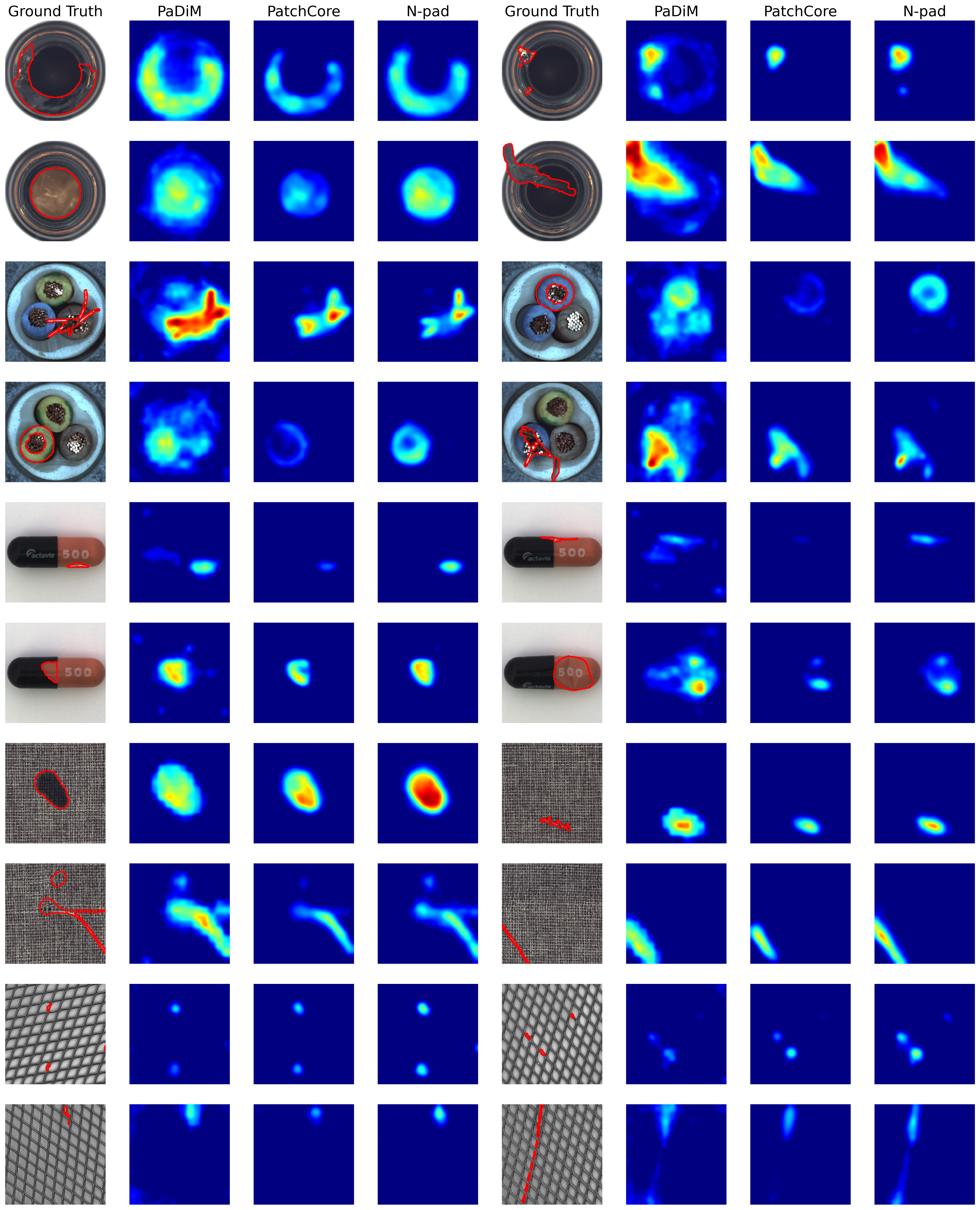}
%\fbox{\rule[-.5cm]{0cm}{4cm} \rule[-.5cm]{4cm}{0cm}}
\end{center}
\caption{Visualization anomalies from top to bottom: bottle, cable, capsule, carpet, and grid.}
\label{img:4}
\end{figure*}

\begin{figure*}[ht]
\begin{center}
%\framebox[4.0in]{$\;$}
\includegraphics[width = 0.95\textwidth]{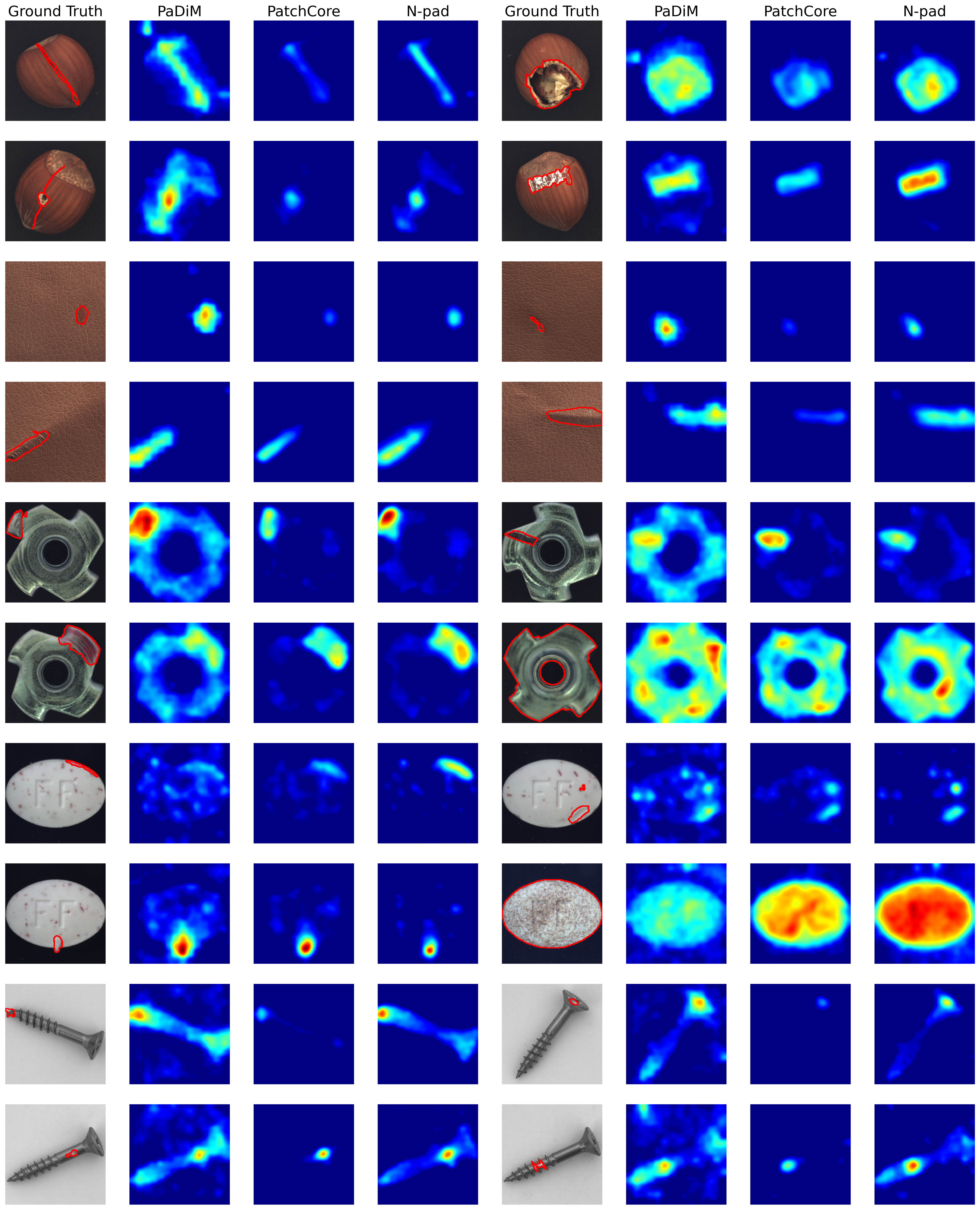}
%\fbox{\rule[-.5cm]{0cm}{4cm} \rule[-.5cm]{4cm}{0cm}}

\end{center}
\caption{Visualization anomalies from top to bottom: hazelnut, leather, metalnut, pill, and screw.}
\label{img:5}
\end{figure*}

\begin{figure*}[ht]
\begin{center}
%\framebox[4.0in]{$\;$}
\includegraphics[width = 0.95\textwidth]{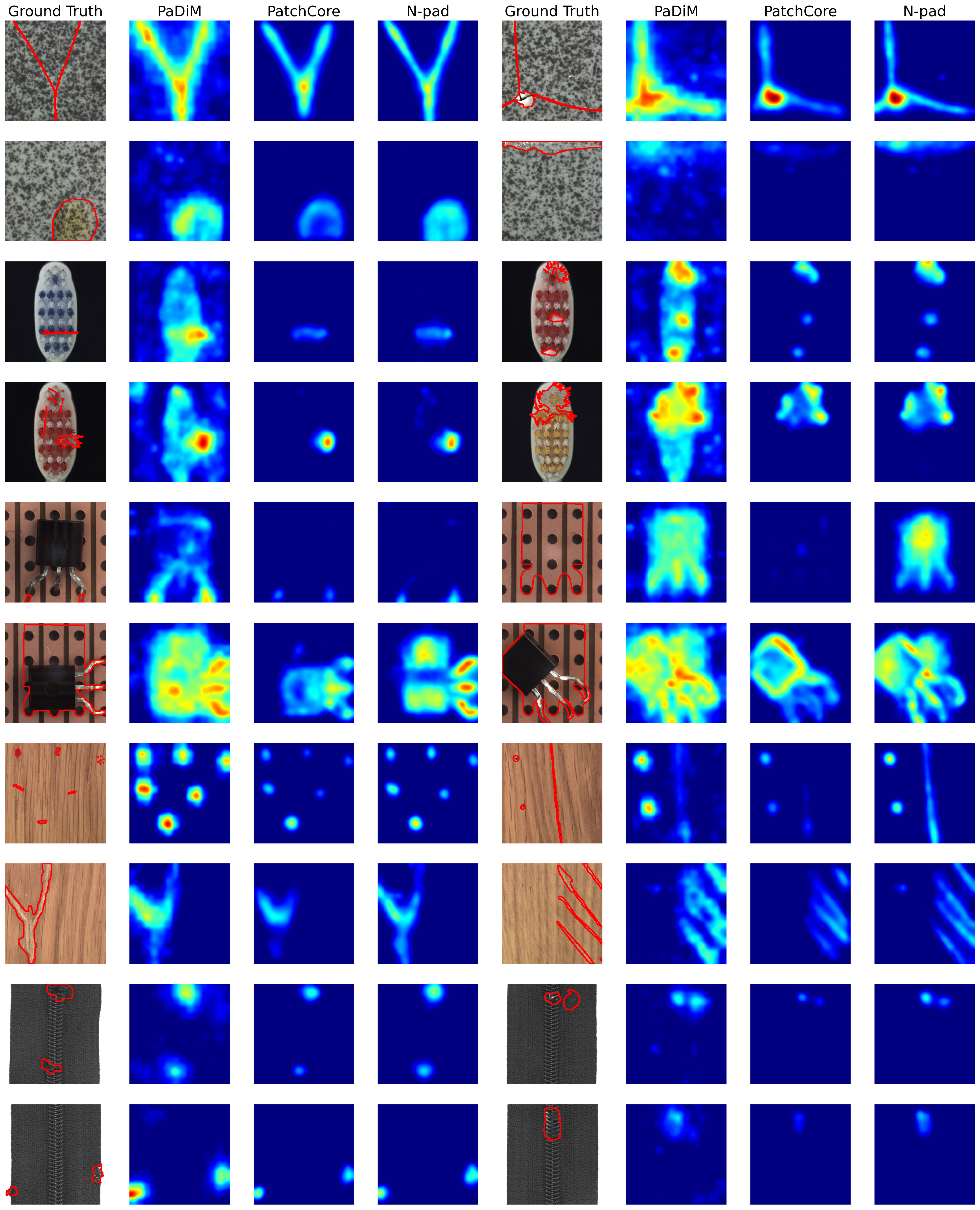}

%\fbox{\rule[-.5cm]{0cm}{4cm} \rule[-.5cm]{4cm}{0cm}}
\end{center}
\caption{Visualization anomalies from top to bottom: tile, toothbrush, transistor, wood and zipper.}
\label{img:6}
\end{figure*}

We present pixel-wise anomaly maps of several anomalous images for all classes in MVTec-AD that were computed by PaDiM, PatchCore and \textit{N-pad} in \cref{img:4,img:5,img:6}. Herein, pixel-wise anomaly maps were visualized by normalizing the anomaly scores from 20\% to 80\% to eliminate relatively nominal pixels and emphasize the anomalous regions. The results are as follows:
\newline
\newline
\newline
\newline
\newline
\newline
\newline
\newline
\newline
\newline
\newline
\newline
\newline
\newline
\newline
\newline
\newline
\newline
\newline
\newline
\newline
\newline
\newline

{\small
\bibliographystyle{ieee_fullname}
\bibliography{egbib}
}

\end{document}

%% file: math_commands.tex
\usepackage{amsmath,amsfonts,bm}

\def\eqref#1{equation~\ref{#1}}
\def\1{\bm{1}}

\def\vmu{{\bm{\mu}}}

\def\mSigma{{\bm{\Sigma}}}

\DeclareMathAlphabet{\mathsfit}{\encodingdefault}{\sfdefault}{m}{sl}
\SetMathAlphabet{\mathsfit}{bold}{\encodingdefault}{\sfdefault}{bx}{n}

\DeclareMathOperator*{\argmax}{arg\,max}